\documentclass[preprint,12pt]{elsarticle}

\usepackage{bm}
\usepackage{amssymb}
\usepackage{amsmath}
\usepackage{amsfonts}
\usepackage{booktabs}
\usepackage{longtable}
\usepackage{setspace}
\usepackage{array}
\usepackage{autobreak}
\usepackage{graphicx}
\usepackage{cite}
\usepackage{url}
\usepackage{breakurl}
\usepackage{subfigure}
\usepackage{float}
\usepackage{bm}
\usepackage[graphicx]{realboxes}
\usepackage{algorithm,algorithmic}

\usepackage {ntheorem}
\newtheorem{assumption}{Assumption}
\newtheorem*{proof}{Proof}
\newtheorem{theorem}{Theorem}
\newtheorem{lemma}[theorem]{Lemma}
\newtheorem{remark}[theorem]{Remark}
\newtheorem{example}{Example}
\usepackage{color}
\usepackage{hyperref}
\hypersetup{hypertex=true,
	colorlinks=true,
	linkcolor=blue,
	anchorcolor=blue,
	citecolor=blue}
\usepackage{caption}
\usepackage[font=small,labelfont=bf,labelsep=none]{caption}
\begin{document}
	\captionsetup[figure]{labelfont={bf},labelformat={default},labelsep=period,name={Fig.}}

\begin{frontmatter}
	
\title{Communication-Efficient $l_0$ Penalized Least Square}
	
\author[1]{Chenqi Gong}
\ead{gcq@stu.cqu.edu.cn}
\author[1]{Hu Yang\corref{cor1}}
\ead{yh@cqu.edu.cn}
\cortext[cor1]{Corresponding author}
\address[1]{College of Mathematics and Statistics, Chongqing University, Chongqing, 401331, China}
\begin{abstract}
In this paper, we propose a communication-efficient penalized regression algorithm for high-dimensional sparse linear regression models with massive data. This approach incorporates an optimized distributed system communication algorithm, named CESDAR algorithm, based on the Enhanced Support Detection and Root finding algorithm. The CESDAR algorithm leverages data distributed across multiple machines to compute and update the active set and introduces the communication-efficient surrogate likelihood framework to approximate the optimal solution for the full sample on the active set, resulting in the avoidance of raw data transmission, which enhances privacy and data security, while significantly improving algorithm execution speed and substantially reducing communication costs. Notably, this approach achieves the same statistical accuracy as the global estimator. Furthermore, this paper explores an extended version of CESDAR and an adaptive version of CESDAR to enhance algorithmic speed and optimize parameter selection, respectively. Simulations and real data benchmarks experiments demonstrate the efficiency and accuracy of the CESDAR algorithm.
\end{abstract}


\end{frontmatter}
\section{Introduction}
\label{sec:1}
The rapid development of data collection techniques has led to unprecedented growth and expansion in both the volume and dimensionality of data. The massive high-dimensional datasets entail high computational costs and memory constraints. Therefore, innovative research in variable selection and parameter estimation methods has become increasingly crucial. Numerous methods have been utilized for variable selection and parameter estimation in the research domain, including LASSO \citep{Tibshirani:1996}, adaptive LASSO \citep{Zou:2006}, the smoothly clipped absolute deviation (SCAD) penalty \citep{Fan:2001}, the minimax concave penalty (MCP) \citep{Zhang:2010} and so on. However, these methods still have some limitations, like necessitating a minimum signal strength for support recovery, the parameter estimation of penalty functions relies on the normalization of the covariate $\bm{X}$ and so on. As a result, $l_0$ regularization has become a popular approach for variable selection. \citet{Huang:2018} proposed a constructive approach for estimating sparse high-dimensional linear regression models, motivated by the computation algorithm of the Karush-Kuhn-Tucker (KKT) conditions for penalized least squares solutions. It iteratively generates a sequence of solutions based on the support using both primal and dual information and root-finding detection. The SDAR algorithm possesses desirable properties and outperforms most existing algorithms in terms of accuracy and efficiency. Next, the Enhanced Support Detection and Root finding method (ESDAR) \citep{Huang:2021} was introduced, which incorporates a step size to balance primitive and dual variables, and the GSDAR fast algorithm was developed for high-dimensional generalized linear models \citep{Huang:2022}. Despite the aforementioned methods being effective in tackling high-dimensional problems, there still exist constraints on the dimensionality $p$ of the samples, demanding that $\log(p)<n$, where $n$ represents the sample size.

With the exponential growth of data volume and increasing concerns regarding privacy, security, machine memory, and communication costs, the development of efficient algorithms to handle massive data instances has become a crucial challenge in statistical research. Constructing a distributed framework is an effective approach to tackle these challenges. Extensive research has been conducted in the field of distributed systems, as well as distributed methods for solving large-scale statistical optimization problems. Several studies, including \citet{Nedic:2009,Johansson:2010,Ram:2010,Agarwal:2011,Dekel:2012,Zhang:2013,Yang:2019}, have explored distributed approaches with notable contributions to computing efficiency. In recent years, numerous scholars have developed a series of efficient methods within the framework of distributed systems. \citet{Jordan:2019} present a communication-efficient surrogate likelihood (CSL) framework for solving distributed statistical inference problems, which provides a communication-efficient surrogate to the global likelihood. Subsequently, the GD-SDAR algorithm proposed by \citet{Wang:2022} and the SD-SDAR algorithm proposed by \citet{Kang:2023} incorporated the SDAR algorithm into a distributed framework, which relax the requirement on $p$ and extends the condition to $\log(p)<N$, where $N=nM$, $n$ represents the sample size on a single machine, $M$ represents the number of machines and $N$ represents the total sample size across all machines.

However, the aforementioned methods suffer from relatively high transmission costs between machines, resulting in diminished algorithmic speed. Moreover, the storage capacity of a single computer is a pivotal requirement. To address these challenges, we introduce a communication-efficient surrogate to the global likelihood function, referred to as the Communication-Efficient Surrogate Likelihood (CSL) \citep{Jordan:2019} framework, building upon the ESDAR algorithm. This framework provides an alternative approach that exhibits high communication efficiency, effectively serving as a substitute for the global likelihood function in the maximum likelihood estimation (MLE) of regular parametric models or the penalized MLE in high-dimensional models.

In this paper, we propose a communication-efficient algorithm called CESDAR, based on the ESDAR algorithm, specifically designed for large-scale data. The CESDAR algorithm simultaneously utilizes multiple machines to work in parallel, each machine performs variable selection and parameter estimation based on locally stored data. Subsequently, the results are transmitted back to the master machine and aggregated through a simple averaging process to obtain the active set. The final result is obtained when the active set no longer changes. When the number of machines $M=1$, the CESDAR algorithm reverts to the ESDAR algorithm. The introduction of multiple machines reduces the running time, diminishes the storage cost of a single computer, and significantly enhances algorithm efficiency. This approach greatly strengthens privacy protection and data transmission security.

The rest of the paper is organized as follows. In Section \ref{sec:2}, we present a comprehensive overview of the CESDAR algorithm, elucidating its intricate methodology. In Section \ref{sec:3}, we give some conditions and theoretical properties of the solution sequence of the CESDAR algorithm. In Sections \ref{sec:4} and \ref{sec:5}, we describe the ECESDAR and ACESDAR algorithms respectively. In Sections \ref{sec:6} and \ref{sec:7}, we provide some simulations and applications. And the summary is given in Section \ref{sec:8}.
\section{CESDAR algorithm}\label{sec:2}
In this section, we first define some symbols as follows for convenience. For the vector $\bm{\beta}=(\beta_1,\ldots,\beta_p)'$, define its $q$-norm by $||\bm{\beta}||_q=\sum\limits_{i=1}^{q}{(|\beta_i|^q)^\frac{1}{q}}$, $q\in[1,\infty]$, and its number of nonzero elements by $||\bm{\beta}||_0$. Let $||\bm{\beta}||_{(T)}$ and $||\bm{\beta}||_{\min}$ be the T-th largest element by magnitude and the minimum absolute value of $\beta$. Denote $\bf0$ as a column vector in $\mathbb{R}^p$ or a matrix whose all elements are 0. Let $|A|$ denote the cardinality of the set A. Denote $\bm{X}_A=(\bm{x}_j,j\in A)\in\mathbb{R}^{N\times|A|}$, where $\bm{x}_j$ is a column of the matrix $\bm{X}$.

We focus on the linear regression model
\begin{equation}
	\label{eq:1}
	\bm{y}=\bm{X\beta}^*+\bm{\varepsilon},
\end{equation}
where $\bm{y}\in \mathbb{R}^N $ is a response vector, $\bm{X}\in\mathbb{R}^{N\times p} $ is the design matrix,  $\bm{\beta}^*=(\beta_1^*,\ldots,\beta_p^*)'$ is the vector of the regression coefficients and $\bm{\varepsilon} \in \mathbb{R}^N $ is a vector of random noise. For data $\lbrace\bm{y},\bm{X}\rbrace$ with a total sample size of $N$, we consider the following objective function\\
\begin{equation}
	\label{eq:2}
	\min\limits_{\beta\in \mathbb{R}^p}{\frac{1}{2N}||\bm{X\beta}-\bm{y}||_2^2}, \enspace subject\enspace to\enspace||\bm{\beta}||_0\le s,
\end{equation}
where $s>0$ controls the sparsity level. 
Next, we consider minimizing the Lagrangian form of the problem (2),\\
\begin{equation}
	\label{eq:3}
	\begin{aligned}
		l_\lambda(\bm{\beta})&={\frac{1}{2N}||\bm{X\beta}-\bm{y}||_2^2+\lambda||\bm{\beta}||_0}\\
		&:=l(\bm{\beta})+\lambda||\bm{\beta}||_0.
	\end{aligned}
\end{equation}
\begin{lemma}
	\label{lemma:1}
	\textit{If $\hat{\bm{\beta}}$ is a global minimizer of (3), then $\hat{\bm{\beta}}$ satisfies\\
		\begin{equation}
			\label{eq:4}
			\left\{
			\begin{aligned}
				\hat{g_i}&=\frac{\partial^2l(\hat{\bm{\beta}}_{-i},\beta_i)}{\partial\beta_i^2}|_{\hat{\beta_i}},&i=1,2,\ldots,p\\
				\hat{d_i}&=-\hat{g_i}^{-1}\frac{\partial l(\hat{\bm{\beta}}_{-i},\beta_i)}{\partial\beta_i}|_{\hat{\beta_i}},&i=1,2,\ldots,p\\
				\hat{\beta_i}&=H_\lambda(\sqrt{\hat{g_i}}|\hat{\beta_i}+\hat{d_i}|),&i=1,2,\ldots,p\\
			\end{aligned}
			\right.
		\end{equation}
		where $l(\hat{\bm{\beta}}_{-i},\beta_i)$ represents the objective function of $\beta_i$ and $H_\lambda(\cdot)$ is defined by
		\begin{equation}
			\label{eq:5}
			H_\lambda(\sqrt{\hat{g_i}}|\hat{\beta_i}+\hat{d_i}|)=\left\{
			\begin{aligned}
				&0,\enspace &if \enspace  \sqrt{\hat{g_i}}|\hat{\beta_i}+\hat{d_i}|<\sqrt{2\lambda}, \\
				&\beta_i,\enspace &if \enspace \sqrt{\hat{g_i}}|\hat{\beta_i}+\hat{d_i}| \ge \sqrt{2\lambda}. \\
			\end{aligned}
			\right.
		\end{equation}
		Conversely, if $\hat{\bm{\beta}}$, $\hat{\bm{d}}$ and $\hat{\bm{g}}$ satisfy (4), then $\hat{\bm{\beta}}$ is a local minimizer of (3).}
\end{lemma}
\begin{remark}
	\label{re:2}
	\textit{Lemma \ref{lemma:1} derives the Karush-Kuhn-Tucker (KKT) condition of the $l_0$  regularized minimization problem (3), which is also derived in \citet{Huang:2018} and \citet{Ming:2024}. There is a proof of Lemma \ref{lemma:1} in \ref{apd:A}.}
\end{remark}

According to Lemma \ref{lemma:1}, we simply define $S=\lbrace1,2,\cdots,p\rbrace$, and respectively denote $supp(\bm{\beta})=\lbrace i\in S|\beta_i\ne 0 \rbrace $ as the support of $\bm{\beta}$. Let $\hat{A}=supp(\hat{\bm{\beta}})$ and $\hat{I}=\hat{A}^c$, then we obtain the elements of the set $\hat{A}$ and $\hat{I}$ as:
$$
\hat{A}=\lbrace i\in S|\sqrt{\hat{g_i}}|\hat{\beta_i}+\hat{d_i}|\ge \sqrt{2\lambda}\rbrace
$$
and
$$
\hat{I}=\lbrace i\in S |\sqrt{\hat{g_i}}|\hat{\beta_i}+\hat{d_i}|< \sqrt{2\lambda}\rbrace.
$$
Then we obtain the equations:
\begin{equation}
	\label{eq:6}
	\left\{
	\begin{aligned}
		&\hat{\bm{\beta}}_{\hat{I}}=\bm{0},\\
		&\hat{\bm{d}}_{\hat{A}}=\bm{0},\\
		&\hat{\bm{\beta}}_{\hat{A}}\in \arg\min l(\bm{\beta}_{\hat{A}}),\\
		&\hat{\bm{g}}=\frac{1}{M}\sum\limits_{m=1}^{M}{\hat{\bm{g}}_m},\\
		&\hat{\bm{d}}=\frac{1}{M}\sum\limits_{m=1}^{M}{\hat{\bm{d}}_m},\\
	\end{aligned}
	\right.
\end{equation}
where $\hat{\bm{g}}_m$ and $\hat{\bm{d}}_m$ respectively represents the vector of $\hat{g}_i$ and $\hat{d}_i$ from
(\ref{eq:4}) for data samples from the dataset of $m$-th client. Furthermore, the estimation of $\hat{\bm{\beta}}_{\hat{A}}$ in (\ref{eq:6}) poses a significant challenge, therefore, we employ the surrogate loss function proposed by \citet{Jordan:2019} to estimate it, that is, 
$$
\tilde{l}(\bm{\beta}_{\hat{A}})=l_1(\tilde{\bm{\beta}}_{\hat{A}})-\langle\bm{\beta}_{\hat{A}},\nabla l_1(\tilde{\bm{\beta}}_{\hat{A}})-\nabla l_N(\tilde{\bm{\beta}}_{\hat{A}})\rangle,
$$
where $\tilde{l}(\bm{\beta}_{\hat{A}})$ is an approximation of $l(\bm{\beta}_{\hat{A}})$, $l_1(\tilde{\bm{\beta}}_{\hat{A}})$ denotes the objective function on the first machine, $\nabla l_1(\tilde{\bm{\beta}}_{\hat{A}})$ represents the gradient on the first machine, $\nabla l_N(\tilde{\bm{\beta}}_{\hat{A}})$ denotes the global gradient, and $\tilde{\bm{\beta}}_{\hat{A}}=\arg\min l_1(\bm{\beta}_{\hat{A}})$. According to \citet{Jordan:2019}, we obtain $\tilde{l}(\bm{\beta}_{\hat{A}})-l(\bm{\beta}_{\hat{A}})=O_p(N^{-\frac{1}{2}})$ obviously.\\

Without loss of generality, we assume that there are $M$ machines, and each machine is allocated $n=N/M$ samples when the total sample size is $N$. For simplicity, we consider the first machine as the master machine. To address the challenges associated with computational time and costs of data transmission and aggregation, we propose a simplified approach in this paper. In our approach, each machine calculates first-order gradient vector $\bm{d}$, second-order derivative vector $\bm{g}$, and $\bm{\beta}$. These vectors are then transmitted to the master machine for simple averaging. The master machine utilizes this information to update the active set and initiates repeated iterations until the active set no longer changes. By distributing the computations of $\bm{d}$, $\bm{g}$, and $\bm{\beta}$ across the machines and performing simple averaging on the master machine, we mitigate the challenges of large-scale data computation and transmission. This enables us to efficiently determine and update the active set, leading to improved optimization results.

Let $\lbrace \bm{\beta}^{(k)}, \bm{d}^{(k)}, \bm{g}^{(k)} \rbrace$ be the $k$-th iteration solution. And we approximate $\lbrace \hat{A}^{(k)}, \hat{I}^{(k)} \rbrace$ by
\begin{equation}
	\label{eq:7}
	\hat{A}^{(k)}=\left\lbrace \sqrt{\hat{\bm{g}}^{(k)}}|\hat{\bm{\beta}}^{(k)}+\hat{\bm{d}}^{(k)}|\ge \sqrt{2\lambda}\right\rbrace,\enspace\hat{I}^{(k)}=(\hat{A}^{(k)})^c.
\end{equation}
The updated $(k+1)$-th iteration solution $\lbrace\bm{\beta}^{(k+1)},\bm{d}^{(k+1)},\bm{g}^{(k+1)}\rbrace$ can be obtained by
\begin{equation}
	\label{eq:8}
	\left\{
	\begin{aligned}
		\hat{\bm{\beta}}_{\hat{I}^{(k)}}^{(k+1)}&=\bm{0}, \\
		\hat{\bm{d}}_{\hat{A}^{(k)}}^{(k+1)}&=\bm{0}, \\
		\hat{\bm{\beta}}_{\hat{A}^{(k)}}^{(k+1)}&\in \arg\min\limits_{\bm{\beta}_{A^{(k)}}}\tilde{l}(\bm{\beta}_{A^{(k)}}),\\
		\hat{\bm{g}}^{(k+1)}&=\frac{1}{M}\sum\limits_{m=1}^{M}{\hat{\bm{g}}_m^{(k+1)}},\\
		\hat{\bm{d}}^{(k+1)}&=\frac{1}{M}\sum\limits_{m=1}^{M}{\hat{\bm{d}}_m^{(k+1)}}.\\
	\end{aligned}
	\right.
\end{equation}
Suppose that our goal is to obtain T-sparse solutions, then we let
\begin{equation}
	\label{eq:9}
	\sqrt{2\lambda^{(k)}}\overset{\bigtriangleup}{=}\left\lVert\sqrt{\hat{\bm{g}}^{(k)}}\cdot|\bm{\beta}^{(k)}+\tau \bm{d}^{(k)}|\right\rVert_{(T)},
\end{equation}
where $\tau\in(0,1]$ is used to balance $\bm{\beta}^{(k)}$ and $\bm{d}^{(k)}$, $T\ge1$ is a given integer. With the initial $\bm{\beta}^{(0)}$, using (\ref{eq:5}) and (\ref{eq:6}) with the $\lambda^{(k)}$ in (\ref{eq:7}), we obtain a sequence of solutions $\lbrace\bm{\beta}^{(k)},k\ge1\rbrace$.
Finally, we transfer the sequence solutions calculated on each machine back to the master machine for simple averaging to get the final sequence solution. We summarize the CESDAR algorithm in Algorithm \ref{alg:1}.
\begin{algorithm}[!ht]
	\caption{Communication efficient support detection and root finding (CESDAR)}
	\label{alg:1}
	\begin{algorithmic}[1] 
		\REQUIRE  Data $\lbrace\bm{y},\bm{X}\rbrace$, $\bm{\beta}^{(0)}$, $\bm{d}^{(0)}$, the number of machines $M$, the integer $T$, the step size $\tau\in(0,1]$; 
		\FOR {$k=0,1,2,\ldots,K$}
		\STATE \begin{small}Set $A^{(k)}=\lbrace i \!\in\! S\mid \sqrt{\bm{g}_i^{(k)}}\lvert\bm{\beta}_i^{(k)}+\tau \bm{d}_i^{(k)}\rvert\ge \lVert\sqrt{\bm{g}^{(k)}}\cdot|\bm{\beta}^{(k)}+\tau\bm{d}^{(k)}|\rVert_{(T)}\rbrace, \; I^{(k)}=(\hat{A}^{(k)})^c$;\end{small}
		\STATE $\bm{\beta}_{I^{(k)}}^{(k+1)}=\bm{0}$;
		\STATE Update $\bm{\beta}_{A^{(k)}}^{(k+1)}\in \arg\min \tilde{l}(\bm{\beta}_{\hat{A}^{(k)}})$;
		\STATE $\bm{g}^{(k+1)}=\frac{1}{M}\sum\limits_{m=1}^{M}{\bm{g}_m^{(k)}}$;
		\STATE $\bm{d}^{(k+1)}=\frac{1}{M}\sum\limits_{m=1}^{M}{\bm{d}_m^{(k)}}$;
		\IF {$A^{(k+1)}=A^{(k)}$}
		\STATE Stop and denote the last iteration by $\bm{\beta}_{\hat{A}}, \bm{\beta}_{\hat{I}}$;
		\ENDIF
		\ENDFOR
		\ENSURE $\hat{\bm{\beta}}=(\bm{\beta}_{\hat{A}}',\bm{\beta}_{\hat{I}}')'$ as the estimate of $\bm{\beta}^*$. 
	\end{algorithmic}
\end{algorithm}
\begin{remark}
	\label{re:3}
	\textit{In Algorithm \ref{alg:1}, we let $\bm{\beta}^{(0)}=\bm{0}$ and terminate the iteration when $A^{(k+1)}=A^{(k)}$. If $k$ is large enough, we have $A^{(k+1)}=A^{(k)}=supp(\bm{\beta}^*)$ under certain conditions.}
\end{remark}
\section{Theoretical Properties}\label{sec:3}
In this section, we establish the $l_2$ and $l_\infty$ error bounds for  $\hat{\bm{\beta}}-\bm{\beta}^*$ and provide the theoretical proof demonstrating that when the target signal exceeds the detectable level, the estimator $\hat{\bm{\beta}}$ achieves the oracle estimator with high probability.

We assume $\bm{X}$ satisfies the sparse Rieze condition (SRC) \citep{Zhang:2008} with order $s$ and spectrum bounds $\lbrace c_-(s),c_+(s)\rbrace$ if 
\begin{align}\nonumber
	0<c_-(s)\le\frac{||\bm{X}_A\bm{u}||^2_2}{N||\bm{u}||^2_2}\le c_+(s)<\infty,\\
	\forall\bm{0}\ne \bm{u}\in\mathbb{R}^{|A|} \enspace with \enspace A\subset S \enspace and \enspace |A|\le s,
	\nonumber
\end{align}
and denote the above condition by $\bm{X}\sim SRC\lbrace s,c_-(s),c_+(s)\rbrace$, which limits the range of the spectrum of the off-diagonal sub-matrices of the Gram matrix $G=X'X/N$. And the spectrum of the off-diagonal sub-matrices of $G$ can be bounded by the sparse orthogonality constants $\theta_{a,b}$, which is defined as follows,
\begin{align}
\theta_{a,b}&\ge \frac{||\bm{X}'_A\bm{X}_B\bm{u}||_2}{N||\bm{u}||_2},\nonumber\\ \forall\bm{0}\not=\bm{u}\in\mathbb{R}^{|B|}\enspace with\enspace A,\; B &\subset S,\enspace|A|\leq a,\enspace |B|\leq b,\enspace and \;A\cap B = \emptyset.
	\nonumber
\end{align}

To obtain distributed conclusions, a series of assumptions need to be introduced. The first assumption delineates the relationship between the parameter space $\Theta$ and the true parameter $\bm{\beta}^*$. The second assumption is the global identifiability condition, which serves as the standard criterion for proving estimation consistency. The final assumption regulates the moments of the high-order derivatives of the loss function, allowing us to obtain estimates of the error with high probability bounds. More details can be found in \citet{Jordan:2019}.
\begin{assumption}
	\label{a:A}
	\textbf{(Parameter space)}\quad The parameter space $\Theta\in \mathbb{R}^p$ is a compact and convex subset, $\bm{\beta}^*\in \Theta$ and $R:=\sup\limits_{\bm{\beta}\in\Theta}||\bm{\beta}-\bm{\beta}^*||_2 > 0$.\\
\end{assumption}
\begin{assumption}
	\label{a:B}
	\textbf{(Identifiability)}\quad For any $\delta>0$, there exists $\epsilon>0$, such that
	\begin{equation}
		\liminf\limits_{n\to\infty}\mathbb{P}\left\lbrace\inf\limits_{||\bm{\beta}-\bm{\beta}^*||_2 > \delta}{(l(\bm{\beta})-l(\bm{\beta}^*))\ge \epsilon}\right\rbrace=1.
		\nonumber
	\end{equation}
\end{assumption}
\begin{assumption}
	\label{a:C}
	\textbf{(Smoothness)}\quad There exist constants $(D,W)$ and a function $G(x)$ such that
	\begin{small}
	\begin{equation}
		\begin{split}
			\mathbb{E}[||\nabla l(\bm{\beta}; X)||_2^{16}]\le D^{16},\;
			\mathbb{E}[||\nabla^2 l(\bm{\beta}; X)-I(\bm{\beta})||_2^{16}]\le W^{16}, \enspace&for\; all\; \bm{\beta}\in U(\rho),\\
			|||\nabla^2l(\bm{\beta};x)-\nabla^2l(\bm{\beta}';x)|||_2\le G(x)||\bm{\beta}-\bm{\beta}'||_2, \qquad\quad&for\; all\; \bm{\beta}, \bm{\beta}'\in U(\rho),
		\end{split}
		\nonumber 
	\end{equation}
	\end{small}where $I(\bm{\beta})=\mathbb{E}(\bm{X}'\bm{X})$,  \begin{small}$|||\nabla^2l(\bm{\beta};x)-\nabla^2l(\bm{\beta}';x)|||_2$\end{small} denotes the spectral norm of \begin{small}$\nabla^2l(\bm{\beta};x)-\nabla^2l(\bm{\beta}';x)$\end{small}, the function $G(x)$ satisfies \begin{small}$\mathbb{E}[G^{16}(X)]\le G^{16}$\end{small} for some constant $G>0$.
\end{assumption}

\subsection{$l_2$ error bounds}
Let $1\le T\le p$ in Algorithm \ref{alg:1}. According to \citet{Huang:2018}, we introduce the following assumptions on design matrix $\bm{X}$ and the error vector $\bm{\varepsilon}$.
\begin{assumption}
	\label{a:1}
	The integer $T$ satisfies $T\ge s$.
\end{assumption}
\begin{assumption}
	\label{a:2}
	Suppose $\bm{X}\!\sim\! SRC\lbrace 2T,c_-(2T),c_+(2T)\rbrace$ for the integer $T$ used in Algorithm \ref{alg:1}.
\end{assumption}
\begin{assumption}
	\label{a:3}
	The independent random errors $\varepsilon_1,\varepsilon_2,\cdots,\varepsilon_n$ are identically distributed with mean zero and sub-Gaussian tails which means there exists a $\sigma\ge 0$ such that $E[\exp(t\varepsilon_i)]\le \exp(\sigma^2t^2/2)$ for $t\in \mathbb{R}^1, i = 1,2,\cdots,n$.
\end{assumption}
\begin{remark}
	\label{re:4}
	\textit{Assumption \ref{a:1} guarantees that the CESDAR algorithm selects a minimum of $J$ significant features. The SRC condition in Assumption \ref{a:2} has been utilized by \citet{Zhang:2008,Zhang:2010,Huang:2018,Kang:2023}. Assumption \ref{a:3} introduces the sub-Gaussian condition, which is commonly assumed in sparse estimation literature and is slightly less demanding than the standard normality assumption. This condition enables the calculation of tail probabilities for specific maximal functions of the error vector $\bm{\varepsilon}$.}
\end{remark}

\begin{theorem}
	\label{th:5}
	Assume Assumptions \ref{a:A}-\ref{a:C} hold, let $T$ be the input integer used in Algorithm \ref{alg:1}, where $1\le T\le p$. $\bm{\beta}^{(k+1)}$ is the solution at $(k+1)$-th iteration.\\
	(i) Assume Assumption \ref{a:1} and Assumption \ref{a:2} hold, we have
	\begin{equation}
		\label{eq:10}
		||\bm{\beta}^{(k+1)}-\bm{\beta}^*||_2\le b_1\gamma^{k}||\bm{\beta}^*||_2+b_2h_2(T),
	\end{equation}
	where
	\begin{equation}
		\begin{split}
			\label{eq:11}
			\gamma=\frac{2\theta_{T,T}+(1+\sqrt{2})\theta_{T,T}^2}{c_-(T)^2}+\frac{(1+\sqrt{2})\theta_{T,T}}{c_-(T)}\;\in\;(0,1),\enspace b_1=1+\frac{\theta_{T,T}}{c_-(T)},\\ b_2=\frac{\gamma}{(1-\gamma)\theta_{T,T}}b_1+\frac{1}{c_-(T)} \enspace and \enspace h_2(T)=\max\limits_{A\subseteq S:|A|\le T}||\bm{X}_A'\bm{\varepsilon}||_2/N.
		\end{split}
	\end{equation}
	(ii) Assume Assumptions \ref{a:1}-\ref{a:3} hold. Then for any $\alpha\in(0,1/2)$, we have
	\begin{equation}
		\label{eq:12}
		||\bm{\beta}^{(k+1)}-\bm{\beta}^*||_2\le b_1\gamma^{k}||\bm{\beta}^*||_2+b_2\eta_1
	\end{equation}
	with probability at least $1-2\alpha$, where
	\begin{equation}
		\label{eq:13}
		\eta_1=\sigma\sqrt{T}\sqrt{2\log(p/\alpha)/N}.
	\end{equation}
	(iii) Assume Assumptions \ref{a:1}-\ref{a:3} hold, then for any $\alpha\in(0,1/2)$, with probability at least $1-2\alpha$, we have
	\begin{equation}
		\label{eq:14}
		||\bm{\beta}^{(k+1)}-\bm{\beta}^*||_2\le c\eta_1, \;if\; k\ge\log_{\frac{1}{\gamma}}\frac{\sqrt{s}||\bm{\beta}^*_{A^*}||_\infty}{\eta_1},
	\end{equation}
	where $c=b_1+b_2$ with $b_1$ and $b_2$ defined in (\ref{eq:11}), and $\eta_1$ is defined in (\ref{eq:13}).
\end{theorem}
\begin{remark}
	\label{re:6}
	Theorem \ref{th:5} provides an upper bound for $l_2$ error, which can reach the theoretical optimum of $O_p(\sqrt{\frac{s\log(p)}{N}})$ after a finite number of iterations, under the assumption that the error term $\bm{\varepsilon}$ follows a sub-Gaussian distribution.
\end{remark}
\begin{theorem}
	\label{th:7}
	Suppose Assumptions \ref{a:A}-\ref{a:3} hold, $||\bm{\beta}^*_{A^*}||_{\min}\ge\frac{\eta_1\gamma}{(1-\gamma)\theta_{T,T}\zeta}$ for some $0<\zeta<1$, then we have
	\begin{equation}
		\label{eq:15}
		A^{(k+1)}\supseteq A^*, \enspace if\; k\ge\log_{\frac{1}{\gamma}}\frac{\sqrt{s}R}{1-\zeta}.
	\end{equation}
	where $R=\frac{||\bm{\beta}^*_{A^*}||_\infty}{||\bm{\beta}^*_{A^*}||_{\min}}$.
\end{theorem}
\begin{remark}
	\label{re:8}
	Theorem \ref{th:7} demonstrates the support of the CESDAR solution sequence covers $A^*$ within a finite number of iterations. Notably, the number of iterations required is $O(\log(\sqrt{s}R))$, depending on the sparsity level $s$ and $R$ denotes the relative magnitude of the coefficients associated with the significant predictors.
\end{remark}
\subsection{$l_\infty$ error bounds}
We replace condition Assumption \ref{a:2} by
\begin{assumption}
	\label{a:4}
	The mutual coherence $\mu$ of $\bm{X}$ satisfies $T\mu\le 1/4$.
\end{assumption}

\begin{theorem}
	\label{th:9}
	Assume Assumptions \ref{a:A}-\ref{a:C} hold, let $T$ be the input integer used in Algorithm \ref{alg:1}, where $1\le T\le p$.\\
	(i) Assume Assumption \ref{a:1} and Assumption \ref{a:2} hold, we have
	\begin{equation}
		\label{eq:16}
		||\bm{\beta}^{(k+1)}-\bm{\beta}^*||_\infty\le \frac{4}{3}\gamma^{k}_\mu||\bm{\beta}^*||_\infty+\frac{4}{3}\left(\frac{4}{1-\gamma_\mu}+1\right)h_\infty(T),
	\end{equation}
	where $\gamma_\mu=\frac{(1+2T\mu)T\mu}{1-(T-1)\mu}+2T\mu$ and $h_\infty(T)=\max\limits_{A\subseteq S:|A|\le T}||\bm{X}_A'\bm{\varepsilon}||_\infty/N$.\\
	(ii) Assume Assumption \ref{a:1}, Assumption \ref{a:3} and Assumption \ref{a:4} hold. Then for any $\alpha\in(0,1/2)$, we have
	\begin{equation}
		\label{eq:17}
		||\bm{\beta}^{(k+1)}-\bm{\beta}^*||_\infty\le \frac{4}{3}\gamma^{k}_\mu||\bm{\beta}^*||_\infty+\frac{4}{3}\left(\frac{4}{1-\gamma_\mu}+1\right)\eta_2,
	\end{equation}
	with probability at least $1-2\alpha$, where
	\begin{equation}
		\label{eq:18}
		\eta_2=\sigma\sqrt{2\log(p/\alpha)/N}.
	\end{equation}
	(iii) Assume Assumption \ref{a:1}, Assumption \ref{a:3} and Assumption \ref{a:4} hold, then for any $\alpha\in(0,1/2)$, with probability at least $1-2\alpha$, we have
	\begin{equation}
		\label{eq:19}
		||\bm{\beta}^{(k+1)}-\bm{\beta}^*||_\infty\le c_\mu\eta_2, \;if\; k\ge\log_{\frac{1}{\gamma_\mu}}\frac{4||\bm{\beta}^*_{A^*}||_\infty}{\eta_2},
	\end{equation}
	where $c_\mu=\frac{16}{3(1-\gamma_\mu)}+\frac{5}{3}$ and $\eta_2$ is defined in (\ref{eq:18}).
\end{theorem}
\begin{remark}
	\label{re:10}
	Theorem \ref{th:9} provides an upper bound for $l_\infty$ error, which can reach the theoretical optimum of $O_p(\sqrt{\frac{\log(p)}{N}})$ after a finite number of iterations, under the assumption that the error term $\bm{\varepsilon}$ follows a sub-Gaussian distribution.
\end{remark}
\begin{theorem}
	\label{th:11}
	Suppose Assumptions \ref{a:A}-\ref{a:3} hold, $||\bm{\beta}^*_{A^*}||_{\min}\ge\frac{4\eta_2}{(1-\gamma_\mu)\zeta}$ for some $0<\zeta<1$, then we have
	\begin{equation}
		A^{(k+1)}\supseteq A^*, \enspace if\; k\ge\log_{\frac{1}{\gamma_\mu}}\frac{R}{1-\zeta}.
	\end{equation}
\end{theorem}
\begin{remark}
	\label{re:12}
	Theorem \ref{th:5} and Theorem \ref{th:7} can be derived from Theorem \ref{th:9} and Theorem \ref{th:11}, respectively, by utilizing the relationship between the $l_\infty$-norm and the $l_2$-norm. We present them separately because the Assumption \ref{a:2} is weaker than Assumption \ref{a:4}. The stronger Assumption \ref{a:4} provides additional insights into CESDAR. Based on this, we can demonstrate that the worst case iteration complexity of CESDAR is independent of the underlying sparsity level.
\end{remark}
\section{Extension}\label{sec:4}
When data is distributed across various regions and machines, the expenses related to data transmission and aggregation can be significant. Inspired by \citet{Ming:2024}, we propose an extended version of the CESDAR algorithm to further reduce communication costs and enhance information security. This extended algorithm is referred to as ECESDAR for simplicity. The ECESDAR algorithm aims to convert the optimization problem, based on data of dimension $N\times p$, into a problem involving transmission data of dimension $n\times T$. In the CESDAR algorithm, communication between machines requires transmitting the vector $\hat{\bm{\beta}}$ (with $T$ non-zero elements and the remaining entries set to zero) as well as the $p$-dimensional vectors $\hat{\bm{d}}$ and $\hat{\bm{g}}$. By contrast, the ECESDAR algorithm necessitates only the transmission of $\hat{\bm{\beta}}$ (with $T$ non-zero elements). This transformation significantly advances information security while simultaneously decreasing both transmission and computation costs. While the ECESDAR algorithm offers enhanced computational efficiency by sacrificing some degree of estimation accuracy. Therefore, this approach is particularly suitable for applications where computational speed is prioritized over precision. We summarized the ECESDAR algorithm in Algorithm \ref{alg:2}. The simulation and application results of the ECESDAR algorithm will be presented in Section \ref{sec:6} and Section \ref{sec:7}.\\
\begin{algorithm}[!ht]
	\caption{The extended version of the CESDAR algorithm}
	\label{alg:2}
	\begin{algorithmic}[1] 
		\REQUIRE  Data $\lbrace \bm{y}^{[1]},\bm{X}^{[1]}\rbrace$ on master machine, $\bm{\beta}^{(0)}$, $\bm{d}^{(0)}$, the number of machines $M$, the integer $T$, the step size $\tau\in(0,1]$;
		\FOR {$k=0,1,2,\ldots,K$}
		\STATE \begin{small}Set $A^{(k)}=\lbrace i\!\in\! S\!\mid\!\sqrt{g_i^{(k)}}|\beta_i^{(k)}+\tau d_i^{(k)}|\ge||\sqrt{\bm{g}^{(k)}}\!\cdot\!|\bm{\beta}^{(k)}+\tau\bm{d}^{(k)}|||_{(T)}\rbrace, \; I^{(k)}=(\hat{A}^{(k)})^c$;\end{small}\\
		\STATE $\bm{\beta}_{I^{(k)}}^{(k+1)}=\bm{0}$;\\
		\STATE Update $\bm{\beta}_{A^{(k)}}^{(k+1)}=\arg\min \tilde{l}(\bm{\beta}_{\hat{A}^{(k)}})$;\\
		\STATE $g_i^{(k+1)}=\frac{\partial^2 l(\bm{\beta})}{\partial\beta_i^2}|_{\beta_i^{(k+1)}},i=1,2,\ldots,p$;\\
		\STATE $d_i^{(k+1)}=-(\hat{g_i^{(k+1)}})^{-1}\frac{\partial l(\bm{\beta})}{\partial\beta_i}|_{\beta_i^{(k+1)}},i\in I^{(k)}$;
		\IF {$A^{(k+1)}=A^{(k)}$}
		\STATE Stop and denote the last iteration by $\bm{\beta}_{\hat{A}}, \bm{\beta}_{\hat{I}}$;
		\ENDIF
		\ENDFOR
		\ENSURE $\hat{\bm{\beta}}=(\bm{\beta}_{\hat{A}}',\bm{\beta}_{\hat{I}}')'$ as the estimate of $\bm{\beta}^*$.
	\end{algorithmic}
\end{algorithm}

\section{Tuning parameter selection}\label{sec:5}
In this section, we present an adaptive version of the CESDAR algorithm that aims to determine the crucial tuning parameter $T$ in Algorithm \ref{alg:1}. This parameter is of great significance as the level of sparsity in the model is often unknown in practical applications. To make an informed decision, various information criteria can be utilized as decision criteria. Examples of such criteria include the composite likelihood version of the Bayesian Information Criterion (BIC) as proposed by \citet{Gao:2010}, the Hierarchical Bayesian Information Criterion (HBIC) introduced by \citet{Wang:2014}, and the Extended Bayesian Information Criterion (EBIC) suggested by \citet{Barber:2015}. In this paper, we specifically consider the following information criterion:
\begin{equation}
	HBIC(\hat{T})=Q(\hat{\bm{\beta}})+\frac{C_N\log{(p)}}{N}|\hat{A}|,
\end{equation}
where $Q(\hat{\bm{\beta}})=\log{(\frac{1}{N}||\bm{y}-\bm{X\hat{\beta}}||_2^2)}$ and $C_N=\log{(\log{(N)})}$.

In adaptive communication efficient support detection and root finding, $T=\hat{T}$ is selected by this criterion and $\hat{\bm{\beta}}(T)$ is used as the final estimate of $\bm{\beta}$. In addition, we take maximum integer $J=\lfloor n/(\log{\log(n)}\cdot\log{(p)})\rfloor$ as in \citet{Fan:2008} and \citet{Huang:2018}. The step size $e$ is the increment of solution path which is usually set to 1, 2, or 4. The summarize ideas is shown in Algorithms \ref{alg:3}.
\begin{algorithm}[!ht]
	\caption{Adaptive CESDAR. (ACESDAR)}
	\label{alg:3}
	\begin{algorithmic}[1] 
		\REQUIRE  Data $\lbrace\bm{y},\bm{X}\rbrace$, $\bm{\beta}^{(0)}$, $\bm{d}^{(0)}$, $\bm{g}^{(0)}$, the number of machines $M$, the step size $e$, an integer $J$;
		\FOR {$l=0,1,2,\ldots,L$}
		\STATE Run Algorithm 1 with tuning parameter $T=el$, data $\lbrace\bm{y}^{[1]},\bm{X}^{[1]}\rbrace$ on the master machine \\and the initial value $(\bm{\beta}^{(l-1)},\bm{d}^{(l-1)},\bm{g}^{(l-1)})$. Denote the output by $\bm{\beta}^{(l)}$;
		\IF {$T>J$}
		\STATE Break;
		\ENDIF
		\STATE Calculate the HBIC value for $\bm{\beta}^{(l)}$, and denote it as HBIC$(T_l)$;
		\ENDFOR
		\STATE Choose the smallest HBIC corresponding to $T$ as the optimal tuning parameter $\hat{T}$;
		\ENSURE $\hat{\bm{\beta}}(\hat{T})$ as the estimate of $\bm{\beta}^*$.
	\end{algorithmic}
\end{algorithm}
\section{Simulation Study}\label{sec:6}
In this section, we present simulation studies to evaluate the performance of our proposed CESDAR and ECESDAR algorithms. Additionally, we compare their performance with that of the ESDAR algorithm. The simulations and application studies were conducted using the R software (version 4.2.0) on a Windows system equipped with the Intel XeonE5-2650 CPU (2.00 GHz, 16 GB RAM). To evaluate the performance of the ESDAR, CESDAR, and ECESDAR algorithms, we utilize the following evaluation metrics and repeat the experiment 100 times.
\begin{itemize}
	\item $AEE=\frac{1}{100}\sum\limits_{a=1}^{100}{||\hat{\bm{\beta}}^{[a]}-\bm{\beta}^*||_2^2}$, \;i.e., average estimation error;
	\item $APE=\frac{1}{100}\sum\limits_{a=1}^{100}{\frac{1}{n_{test}}||\bm{X}_{test}^{[a]}\hat{\bm{\beta}}^{[a]}-\bm{y}_{test}^{[a]}||_2^2}$, \;i.e., average prediction error;
	\item $APDR=\frac{1}{100}\sum\limits_{a=1}^{100}{(\frac{|\hat{A}\cap A^*|}{|A^*|})^{[a]}}$, \;i.e., average positive discovery rate;
	\item $AFDR=\frac{1}{100}\sum\limits_{a=1}^{100}{(\frac{|\hat{I}\cap I^*|}{|I^*|})^{[a]}}$, \; i.e., average false discovery rate,
\end{itemize}
where $[a]$ represents the results of the $a$-th experiment.\\

In the simulation study, we hope that AEE and APE of the model to be as close as possible to 0, and APDR and AFDR to be as close as possible to 1. In addition, we also recorded the rate at which the Oracle estimator is achieved by the CESDAR algorithm over 100 replicates (ORA), the average number of iterations (ANI) and the average running time that ignoring the time spent transferring data between machines (ART).
\subsection{Simulation Examples}
We generate the design matrix $\bm{X}$ as $\bm{X}_i\sim N(0,I_{p})$, $i=1,2,\ldots,n$, and set $\beta_j^*\sim \textbf{Uniform}(r_*,r^*)$, $j\in A^*$ and $\bm{\beta}_{I^*}^*=0$ , where $A^*$ is a random chosen subset of $S=\lbrace1,2,\ldots,p\rbrace$. We fix $\tau=0.5$, $r_*=\sqrt{2log(p)/N}$, $r^*=Rr_*$, where $R=20$.
\begin{example}
	\label{ex:1}
	In this example, we investigate the effect of the number of machines $M$ on the model estimate when $N>p$. We generate $\bm{X}$ with $N=100,000$, $p=500$, $s=10$. Here, we fix $T=s$. The number of machines $M$ are set to 2, 4, 8, 16, 32, 64 and 128. The simulation results are shown in Table \ref{tab:1}.
\end{example}
\begin{example}
	\label{ex:2}
	In this example, we investigate the effect of the number of machines $M$ on the model estimate when $N<p$. We generate $\bm{X}$ with $N=5,000$, $p=10,000$, $s=10$. Here, we fix $T=s$. The number of machines $M$ are set to 2, 4, 6, 8, 10, 12, 14 and 16. The simulation results are shown in Table \ref{tab:2}.
\end{example}
\begin{example}
	\label{ex:3}
	In this example, we study the influence of the dimension $p$ for high-dimensional sparse regression model with large-scale data. We generate $\bm{X}$ with $N=5,000$, $s=10$ and let $M=5$. The dimension $p$ upgrades from 2,000 to 10,000 with a step size of 2000. $T$ is taken from 2 to 20 with a step of 2. The simulation results are shown in Figure \ref{fig:1}. In particular, CESDAR(Machine 1) means we only assigned this machine 10,000 samples to implement CESDAR algorithm, which is equivalent to the computation of one of the machines when $M=5$.
\end{example}
\begin{example}
	\label{ex:4}
	In this example, we study the influence of sparsity level $s$ for high-dimensional sparse regression model with large-scale data. We generate $\bm{X}$ and set $N=5,000$, $p=10,000$, $M=5$. The sparsity level $s$ upgrades from 2 to 20 with a step size of 2. $T$ is taken from 2 to 20 with a step of 2. The simulation results are shown in Figure \ref{fig:2}.
\end{example}
\subsection{Simulation results and analysis}
\begin{figure}[!htbp]
	\centering 
	\includegraphics[height=15cm,width=15cm]{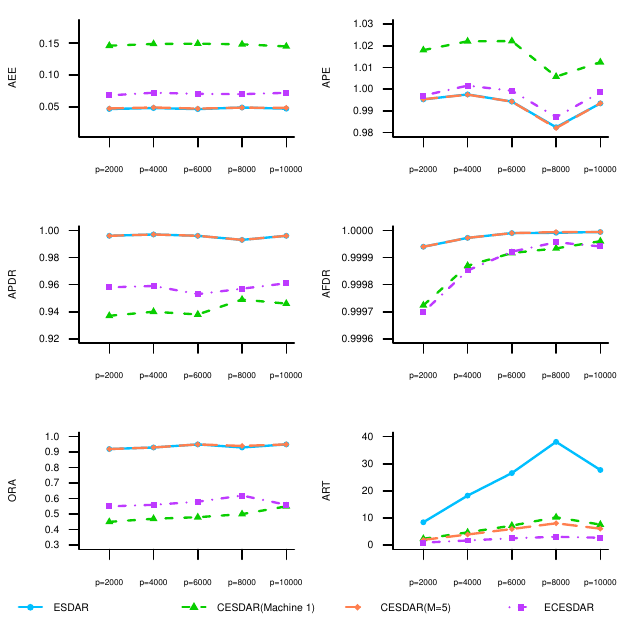}
	\caption{Simulation results for ESDAR, CESDAR and ECESDAR in Example \ref{ex:3}.}
	\label{fig:1}
\end{figure}

\begin{figure}[!htbp]
	\centering 
	\includegraphics[height=15cm,width=15cm]{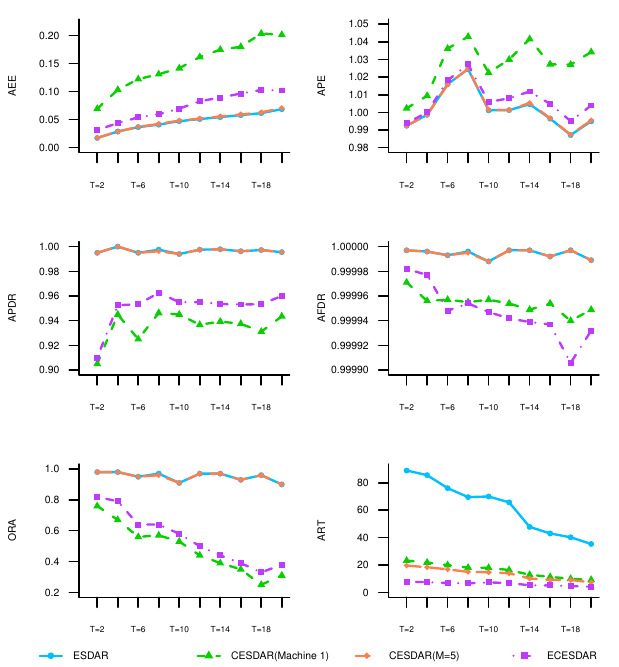}
	\caption{Simulation results for ESDAR, CESDAR and ECESDAR in Example \ref{ex:4}.}
	\label{fig:2}
\end{figure}

In Tables \ref{tab:1}-\ref{tab:2} and Figures \ref{fig:1}-\ref{fig:2}, we can make the following observations.

Table \ref{tab:1} demonstrates the results of Example \ref{ex:1}, illustrating that the performance of the ESDAR, CESDAR and ECESDAR algorithms, as measured by AEE, APE, APDR, AFDR, and ORA, exhibits a slight decline as the number of machines increases when $N>p$. Additionally, the time requires by the CESDAR and ECESDAR algorithms decreases significantly as the number of machines increases. This highlights the essential nature of collaborative endeavors among machines in scenarios characterized by large volumes of data, urgent result requirements, and low economic costs. Notably, the loss incurred by the ECESDAR algorithm is slightly greater than that of the CESDAR algorithm, owing to the former's emphasis on algorithmic speed improvement, rendering it suitable for relatively urgent tasks.

Table \ref{tab:2} illustrates the results of Example \ref{ex:2}, indicating that the performance of ESDAR, CESDAR, and ECESDAR algorithms shows a decline in terms of AEE, APE, APDR, AFDR, and ORA with the increase of machines, when $N<p$. Furthermore, when the number of samples is limited, dividing these among the machines may lead to marginally larger errors as compared to experimental results derived from a larger number of samples. It's worth noting that the time consumption of the CESDAR and ECESDAR algorithms significantly decreases with an increase in the number of machines.\\
\begin{table}[!htbp]
	\caption{Simulation results for Example \ref{ex:1} with different number of machines $M$.} 
	\label{tab:1}
	\centering
	\resizebox{1.0\linewidth}{!}{
		\begin{tabular}{ccccccccc} 
			\toprule 
			Number of Machines&Method&AEE&APE&APDR&AFDR&ORA&ANI&ART\\
			\midrule 
			M=1&ESDAR&0.00996(0.00310)&1.00685(0.13765)&0.994&0.99988&0.94&1.07&0.7814\\
			\multicolumn{1}{c}{M=2}&CESDAR&0.00996(0.00310)&1.00685(0.13765)&0.994&0.99988&0.94&1.07&0.4103\\
			&ECESDAR&0.01164(0.00457)&1.00721(0.13732)&0.976&0.99951&0.77&1.19&0.1757\\
			\multicolumn{1}{c}{M=4}&CESDAR&0.00995(0.00310)&1.00686(0.13766)&0.994&0.99988&0.94&1.07&0.2011\\
			&ECESDAR&0.01322(0.00557)&1.00714(0.13748)&0.957&0.99912&0.62&1.46&0.1016\\
			\multicolumn{1}{c}{M=8}&CESDAR&0.01001(0.00311)&1.00687(0.13765)&0.994&0.99988&0.94&1.07&0.1031\\
			&ECESDAR&0.02011(0.01079)&1.00773(0.13758)&0.918&0.99833&0.36&1.87&0.0606\\
			\multicolumn{1}{c}{M=16}&CESDAR&0.01013(0.00298)&1.00685(0.13761)&0.994&0.99988&0.94&1.07&0.0512\\
			&ECESDAR&0.03080(0.01602)&1.00828(0.13713)&0.867&0.99729&0.19&1.62&0.0284\\
			\multicolumn{1}{c}{M=32}&CESDAR&0.01071(0.00308)&1.00689(0.13768)&0.994&0.99988&0.94&1.07&0.0263\\
			&ECESDAR&0.04944(0.02283)&1.01031(0.13808)&0.808&0.99608&0.07&2.59&0.0232\\
			\multicolumn{1}{c}{M=64}&CESDAR&0.01221(0.00326)&1.00689(0.13766)&0.994&0.99988&0.94&1.07&0.0137\\
			&ECESDAR&0.08035(0.02875)&1.01675(0.13869)&0.715&0.99418&0.01&4.92&0.0226\\
			\multicolumn{1}{c}{M=128}&CESDAR&0.01674(0.00512)&1.00735(0.13809)&0.994&0.99988&0.94&1.07&0.0079\\
			&ECESDAR&0.13448(0.04073)&1.02375(0.13844)&0.597&0.99178&0.00&7.00&0.0168\\
			\bottomrule 
	\end{tabular}}
\end{table}
\begin{table}[!htbp]
	\caption{Simulation results for Example \ref{ex:2} with different number of machines $M$.}
	\label{tab:2}
	\centering
	\resizebox{1.0\linewidth}{!}{
		\begin{tabular}{ccccccccc} 
			\toprule 
			Number of Machines&Method&AEE&APE&APDR&AFDR&ORA&ANI&ART\\
			\midrule 
			M=1&ESDAR&0.04591(0.01121)&0.99770(0.13352)&0.998&1.00000&0.98&1.47&0.8846\\
			\multicolumn{1}{c}{M=2}&CESDAR&0.04600(0.01123)&0.99767(0.13344)&0.998&1.00000&0.98&1.47&0.4568\\
			&ECESDAR&0.05369(0.02481)&0.99922(0.13316)&0.984&0.99998&0.86&1.74&0.1756\\
			\multicolumn{1}{c}{M=4}&CESDAR&0.04625(0.01117)&0.99783(0.13342)&0.998&1.00000&0.98&1.47&0.2305\\
			&ECESDAR&0.06886(0.03936)&1.00094(0.13401)&0.963&0.99996&0.67&2.18&0.1030\\
			\multicolumn{1}{c}{M=6}&CESDAR&0.04719(0.01128)&0.99818(0.13340)&0.998&1.00000&0.98&1.47&0.1556\\
			&ECESDAR&0.09017(0.05498)&1.00577(0.13564)&0.940&0.99994&0.53&3.58&0.1018\\
			\multicolumn{1}{c}{M=8}&CESDAR&0.04864(0.01126)&0.99845(0.13319)&0.998&1.00000&0.98&1.47&0.1183\\
			&ECESDAR&0.10273(0.06191)&1.00879(0.13539)&0.927&0.99993&0.46&4.14&0.0879\\
			\multicolumn{1}{c}{M=10}&CESDAR&0.05044(0.01191)&0.99865(0.13306)&0.998&1.00000&0.98&1.47&0.0957\\
			&ECESDAR&0.13719(0.07403)&1.01969(0.13736)&0.898&0.99990&0.32&6.06&0.0993\\
			\multicolumn{1}{c}{M=12}&CESDAR&0.05221(0.01285)&0.99910(0.13344)&0.998&1.00000&0.98&1.47&0.0809\\
			&ECESDAR&0.15557(0.08380)&1.02951(0.14444)&0.887&0.99989&0.27&6.52&0.0880\\
			\multicolumn{1}{c}{M=14}&CESDAR&0.05562(0.01388)&0.99967(0.13278)&0.998&1.00000&0.98&1.47&0.0705\\
			&ECESDAR&0.17685(0.09239)&1.04054(0.15114)&0.870&0.99987&0.23&7.11&0.0803\\
			\multicolumn{1}{c}{M=16}&CESDAR&0.05837(0.01496)&0.99971(0.13255)&0.998&1.00000&0.98&1.47&0.0628\\
			&ECESDAR&0.19190(0.09564)&1.04346(0.14577)&0.859&0.99986&0.21&8.28&0.0836\\
			\bottomrule 
	\end{tabular}}
\end{table}
Figure \ref{fig:1} depicts that the estimates generated by the CESDAR and ECESDAR algorithms are relatively insensitive to the data dimension. However, comparing it with the ESDAR algorithm, it is evident that the CESDAR and ECESDAR algorithms exhibit significantly reduced computation time as the data dimension increases. This observation implies that when dealing with high-dimensional data, the CESDAR and ECESDAR algorithms can be employed to substantially decrease the running time while making only negligible sacrifices in accuracy. Notably, the ECESDAR algorithm exhibits slightly inferior performance compared to the CESDAR algorithm, but it compensates by reducing runtime, as well as decreasing computational and transmission costs. Furthermore, CESDAR (M=5) and ESDAR show virtually identical performance in all aspects except for time, which further underscores how the CESDAR algorithm achieves faster runtime while maintaining a minimal error rate.

Figure \ref{fig:2} demonstrates that the estimates obtained from the CESDAR and ECESDAR algorithms are greatly impacted by the sparsity levels in the data. Remarkably, despite significantly reducing computation time, the CESDAR and ECESDAR algorithms guarantee that the resulting estimates remain within a narrow error margin. Thus, the CESDAR and ECESDAR algorithms are well-suited for large-scale data computations. Similar to Example \ref{ex:3}, CESDAR (M=5) and ESDAR exhibit nearly indistinguishable performance across all facets, with the exception of time, thereby reinforcing the notion that the CESDAR algorithm attains expedited runtime while upholding a negligible error rate.
\section{Real Data Applications}\label{sec:7}
In this section, we apply the proposed CESDAR and ECESDAR algorithms on two real data sets: Avocado price data and Human resource data.
\subsection{Avocado Price Data}
To illustrate the superiority of CESDAR and ECESDAR algorithms, we first apply it to avocado price data. The data set can be obtained at \url{https://aistudio.baidu.com/aistudio/datasetdetail/109315}. The data set is a collection of 33,045 pieces of historical data on avocado prices and sales in multiple cities and states from 2015 to 2020. The task is to predict the average price of a single avocado based on region, year, type of avocado (conventional or organic), total number of avocados sold, and the number of avocados sold in three major varieties coded as 4046, 4225, and 4770. The data contains 5 continuous variables and 2 categorical variables, where the categorical variables are transformed into dummy variables of appropriate size. In addition, to verify the variable selection ability of the algorithm, we expanded the data dimension, randomly generated a 1993 dimensional redundant variable with a mean of 0 and a variance of 1, and merged it with the original data. The combined data had a total of 2000 dimensions, and the 2000-dimensional variables are coded as NO.1-NO.2000 respectively. All numerical variables are normalized to mean zero and variance one. The complete variable description is shown in Table \ref{tab:3}. In order to compare the performance of different methods and obtain variable importance ranking, we randomly selected 30,000 samples as the training set and the remaining 3,045 samples as the test set. The process was repeated 500 times, and the simulation results were shown in Table \ref{tab:4} and Table \ref{tab:5}.
\begin{table}[!htbp]
	\caption{Variable description for avocado price data. Non-categorical numerical variables are standardized to have mean zero and variance one.} 
	\label{tab:3}
	\centering
	\resizebox{1.0\linewidth}{!}{
		\begin{tabular}{ccc} 
			\toprule 
			Variable&Description&Variable used in the model\\
			\midrule 
			AveragePrice&The average price of a single avocado&Used as the response variable\\
			Total Volume (NO.1)&Total number of avocados sold&Used as numerical variable\\
			4046 (NO.2)&Number of avocados coded 4046 sold&Used as numerical variable\\
			4225 (NO.3)&Number of avocados coded 4225 sold&Used as numerical variable\\
			4770 (NO.4)&Number of avocados coded 4770 sold&Used as numerical variable\\
			Year(NO.5)&Year between 2015 and 2020&Used as numerical variable\\
			Type (NO.6)&conventional or organic&Converted to 2 dummies\\
			Region (NO.7)&Observed city or region&Converted to 11 dummies\\
			\bottomrule 
	\end{tabular}}
\end{table}
\begin{table}[!htbp]
	\caption{Variable selection results and frequency of top 7 variables selected of ESDAR, CESDAR (Machine 1 and M=5) and ECESDAR algorithms when sparsity level is set to 7 (same as real sparse level). When CESDAR(Machine 1) is run, only 1/5 of the sample size of the training set is assigned to this machine, that is, 10,000 samples, which is equivalent to the result of one of the machines running CESDAR(M=5).} 
	\label{tab:4}
	\centering
	\resizebox{1.0\linewidth}{!}{
		\begin{tabular}{cccccccc} 
			\toprule 
			\multicolumn{2}{c}{ESDAR}&\multicolumn{2}{c}{CESDAR(Machine 1)}&\multicolumn{2}{c}{CESDAR(M=5)}&\multicolumn{2}{c}{ECESDAR}\\
			\cmidrule(lr){1-2}\cmidrule(lr){3-4}\cmidrule(lr){5-6}\cmidrule(lr){7-8}
			Selected Variables&Frequency&Selected Variables&Frequency&Selected Variables&Frequency&Selected Variables&Frequency\\
			\cmidrule(lr){1-2}\cmidrule(lr){3-4}\cmidrule(lr){5-6}\cmidrule(lr){7-8}
			NO.1&500&NO.1&500&NO.1&500&NO.1&500\\
			NO.2&500&NO.2&500&NO.2&500&NO.2&500\\
			NO.3&500&NO.3&500&NO.3&500&NO.3&500\\
			NO.4&500&NO.6&500&NO.4&500&NO.4&500\\
			NO.6&500&NO.4&490&NO.6&500&NO.6&500\\
			NO.80&499&NO.80&13&NO.80&499&NO.80&13\\
			NO.562&499&NO.1043&12&NO.562&477&NO.562&12\\
			\bottomrule 
	\end{tabular}}
\end{table}
\begin{table}[!htbp]
	\caption{The average prediction error (APE) and its variance, the average number of iterations (ANI) and the average running time (ART) of 500 repeated experiments for ESDAR, CESDAR(Machine 1), CESDAR(M=5) and ECESDAR.}
	\label{tab:5}
	\centering
	\begin{tabular}{cccc} 
		\toprule 
		Method&APE&ANI&ART\\
		\midrule 
		ESDAR&0.0605(0.0012)&2.00&1.4108\\
		CESDAR(Machine 1)&0.0608(0.0015)&3.03&0.3570\\
		CESDAR(M=5)&0.0604(0.0012)&2.00&0.2812\\
		ECESDAR&0.0604(0.0012)&2.36&0.1165\\
		\bottomrule 
	\end{tabular}
\end{table}

Table \ref{tab:4} shows the importance ranking of variables obtained in 500 replicates of ESDAR, CESDAR(Machine 1), CESDAR(M=5) and ECESDAR. It can be seen that the first 5 variables selected by the three include variables NO.1, NO.2, NO.3, NO.4, and NO.6, and the selection rate of the above 5 major variables exceeds 98.00\%. Compared with CESDAR(Machine 1), CESDAR(M=5) can more clearly select major variable NO.4. Table \ref{tab:5} illustrates the average prediction error and its variance, the average number of iterations and the average running time of 500 repeated experiments of ESDAR, CESDAR(Machine 1), CESDAR(M=5) and ECESDAR. It can be seen from the above results that CESDAR and ECESDAR algorithms can greatly reduce the running time and the number of iterations while ensuring the accuracy of model estimation results as much as possible. From the above results, it is evident that CESDAR(M=5), when compared to ESDAR, not only achieves similar variable selection performance but also significantly reduces runtime. This approach ensures high estimation accuracy while greatly saving computational costs and minimizing memory consumption.
\subsection{Human Resource Data}
In this subsection, we applied the CESDAR and ECESDAR algorithms to analyze human resource data. The dataset used in this study can be obtained at \url{https://aistudio.baidu.com/datasetdetail/166181}. It consists of 15,000 records related to employee work in a company. The goal of this task is to predict employee salaries based on various factors, including satisfaction data analysis, last evaluation, number of projects, number of hours worked per month, years worked, accident rate, turnover, promotion rate within five years, and department. The dataset contains eight continuous variables and one categorical variable, which was transformed into dummy variables of appropriate size. To assess the algorithm's variable selection capability, we introduced additional variables to the dataset. These variables were randomly generated and had 1,991 dimensions, with a mean of 0 and a variance of 1. The expanded dataset had a total of 2,000 dimensions, where the 2,000-dimensional variables were labeled as NO.1-NO.2000. Table \ref{tab:6} provides a complete description of the variables. To compare the performance of different methods and rank the importance of variables, we randomly selected 14,000 samples as the training set and reserved the remaining 1,000 samples as the test set. This process was repeated 500 times, and the simulation results are presented in Tables \ref{tab:7} and \ref{tab:8}.
\begin{table}[!htbp]
	\caption{Variable description for human resource data.} 
	\label{tab:6}
	\centering
	\resizebox{1.0\linewidth}{!}{
		\begin{tabular}{ccc} 
			\toprule 
			Variable&Description&Variable used in the model\\
			\midrule 
			Salary&Employee salary level&Used as the response variable\\
			SL (NO.1)&Satisfaction level&Used as numerical variable\\
			LE (NO.2)&Last evaluation&Used as numerical variable\\
			NP (NO.3)&Number of projects&Used as numerical variable\\
			NH (NO.4)&Number of hours worked per month&Used as numerical variable\\
			WY (NO.5)&Working years&Used as numerical variable\\
			AS (NO.6)&Accident situation&Used as numerical variable\\
			TS (NO.7)&Turnover situation&Used as numerical variable\\
			PR (NO.8)&Promotion rate in five years&Used as numerical variable\\
			DEP (NO.9)&Department&Converted to 10 dummies\\
			\bottomrule 
	\end{tabular}}
\end{table}
\begin{table}[!htbp]
	\caption{Variable selection results and frequency of top 6 variables selected of ESDAR, CESDAR (Machine 1 and M=5) and ECESDAR algorithms when sparsity level is set to 6.} 
	\label{tab:7}
	\centering
	\resizebox{1.0\linewidth}{!}{
		\begin{tabular}{cccccccc} 
			\toprule 
			\multicolumn{2}{c}{ESDAR}&\multicolumn{2}{c}{CESDAR(Machine 1)}&\multicolumn{2}{c}{CESDAR(M=5)}&\multicolumn{2}{c}{ECESDAR}\\
			\cmidrule(lr){1-2}\cmidrule(lr){3-4}\cmidrule(lr){5-6}\cmidrule(lr){7-8}
			Selected Variables&Frequency&Selected Variables&Frequency&Selected Variables&Frequency&Selected Variables&Frequency\\
			\cmidrule(lr){1-2}\cmidrule(lr){3-4}\cmidrule(lr){5-6}\cmidrule(lr){7-8}
			NO.7&500&NO.7&500&NO.7&500&NO.7&500\\
			NO.8&500&NO.8&424&NO.8&500&NO.8&428\\
			NO.60&500&NO.715&40&NO.715&500&NO.715&40\\
			NO.715&500&NO.1936&14&NO.1911&447&NO.5&18\\
			NO.5&499&NO.60&13&NO.5&305&NO.1911&18\\
			NO.1911&499&NO.1911&13&NO.9&165&NO.1936&18\\
			\bottomrule 
	\end{tabular}}
\end{table}
\begin{table}[!htbp]
	\caption{The average prediction error (APE) and its variance, the average number of iterations (ANI) and the average running time (ART) of 500 repeated experiments for ESDAR, CESDAR(Machine 1), CESDAR(M=5) and ECESDAR.} 
	\label{tab:8}
	\centering
	\begin{tabular}{cccc} 
		\toprule 
		Method&APE&ANI&ART\\
		\midrule 
		ESDAR&0.3273(0.0022)&2.00&0.6931\\
		CESDAR(Machine 1)&0.3366(0.0065)&4.00&0.2047\\
		CESDAR(M=5)&0.3264(0.0026)&1.67&0.1180\\
		ECESDAR&0.3305(0.0052)&2.75&0.0621\\
		\bottomrule 
	\end{tabular}
\end{table}

Table \ref{tab:7} displays the importance ranking of variables obtained from 500 replicates of the ESDAR algorithm, CESDAR (Machine 1), and CESDAR (M=5). It can be observed that the top two important variables selected by all three algorithms are variables NO.7 and NO.8. Table \ref{tab:8} presents the average prediction error and its variance, average number of iterations, and average running time across the 500 repeated experiments of the ESDAR algorithm, CESDAR (Machine 1), CESDAR (M=5) and ECESDAR. The results demonstrate that the CESDAR (M=5) and ECESDAR algorithms significantly reduces computation time while achieving a minimized average prediction error.
\section{Concluding Remarks}\label{sec:8}
In this paper, we present a novel CESDAR algorithm designed specifically for large-scale data. Our algorithm focuses on achieving high communication efficiency while fitting sparse, high-dimensional linear regression models. This distributed computational method extends the computable data size and dimension based on the ESDAR algorithm, ensuring efficient estimation while reducing computation and communication costs. We conducted a simulation study to validate the accuracy and efficiency of the CESDAR algorithms. The results demonstrate that the CESDAR algorithm achieves faster running speeds compared to the ESDAR algorithm while maintaining result accuracy, especially when dealing with rapidly expanding data sizes and dimensions. Furthermore, the CESDAR algorithm offers significant reductions in machine storage costs.

\section*{Declaration of competing interest}
The authors declare that they have no known competing financial interests or personal relationships that could have appeared to
influence the work reported in this paper.

\section*{Data availability}
Data will be made available on request.

\section*{Acknowledges}
{This work is supported by the National Natural Science Foundation of China [Grant No.12371281].}
\newpage

\appendix
\section{}
\label{apd:A}
\setcounter{equation}{0}
\renewcommand\theequation{A.\arabic{equation}}
\noindent\textbf{Proof of Lemma \ref{lemma:1}.}
\begin{proof}
	As Huang et al. (2018), let $l_\lambda(\bm{\beta})=\frac{1}{2N}||\bm{X\beta}-\bm{y}||^2_2+\lambda||\bm{\beta}||_0$. Assume $\hat{\bm{\beta}}$ is the global minimizer of $l_\lambda$, then we have
	\begin{equation}
		\begin{split}
			\label{A1}
			&\hat{\beta}_i\in \arg\min\limits_{\phi\in \mathbb{R}^p}l_\lambda(\hat{\beta_1},\ldots,\hat{\beta}_{i-1},\phi,\hat{\beta}_{i+1},\ldots,\hat{\beta}_p)\\ 
			&\Rightarrow\hat{\beta}_i\in \arg\min\limits_{\phi\in \mathbb{R}^p}\frac{1}{2N}||\bm{X}\hat{\bm{\beta}}+(\phi-\hat{\beta}_i)X_i-\bm{y}||^2_2+\lambda|\phi|_0\\
			&\Rightarrow\hat{\beta}_i\in \arg\min\limits_{\phi\in \mathbb{R}^p}\frac{\bm{X}'_i\bm{X}_i}{2N}(\phi-\hat{\beta}_i)^2+\frac{1}{N}(\phi-\hat{\beta}_i)\bm{X}_i'(\bm{X}\hat{\bm{\beta}}-\bm{y})+\lambda|\phi|_0\\
			&\Rightarrow\hat{\beta}_i\in \arg\min\limits_{\phi\in \mathbb{R}^p}\frac{\bm{X}'_i\bm{X}_i}{2N}(\phi-(\hat{\beta}_i+\bm{X}'_i(\bm{y}-\bm{X}\hat{\bm{\beta}})/N))^2+\lambda|\phi|_0,
		\end{split}
	\end{equation}
	where $\bm{X}'_i\bm{X}_i=\hat{g_i}$.
	According to the defined $H_\lambda(\cdot)$ in (\ref{eq:5}), we obtain $\hat{\beta}_i=H_\lambda(\sqrt{\hat{g_i}  }|\hat{\beta}_i+\hat{d}_i|)$, $i=1,\cdots,p$, (\ref{eq:4}) holds as a consequence.
	
	We rewrite the loss function in the distributed way, i.e.
	\begin{equation}
		\begin{split}
			\label{A2}
			l_\lambda(\bm{\beta})&=\frac{1}{2N}||\bm{X\beta}-\bm{y}||^2_2+\lambda||\bm{\beta}||_0\\
			&=\frac{1}{2N}\sum_{m=1}^M||\bm{X}_m\bm{\beta}-\bm{y}_m||^2_2+\lambda||\bm{\beta}||_0\\
			&=\frac{1}{M}\sum_{m=1}^M\frac{1}{2n}||\bm{X}_m\bm{\beta}-\bm{y}_m||^2_2+\lambda||\bm{\beta}||_0.
		\end{split}
	\end{equation}
	Summarizing the above conclusions we get $\hat{d_i}=-\hat{g_i}^{-1}\frac{\partial l(\hat{\bm{\beta}}_{-i},\beta_i)}{\partial\beta_i}|_{\hat{\beta_i}}$, $i=1,2,\ldots,p$, and the sufficiency of Lemma \ref{lemma:1} can be proved.
	
	Then we prove the necessity of Lemma \ref{lemma:1}. Suppose (4) holds, and we let $\hat{A}=\lbrace i \in S \mid \sqrt{\hat{g}_i}|\hat{\beta}_i+\hat{d}_i|\ge\sqrt{2\lambda}\rbrace$. According to the definition of $H_\lambda(\cdot)$, we show that for $i\in \hat{A}$, $|\hat{\beta}_i|\ge\sqrt{2\lambda}$. Furthermore, $\hat{\bm{d}}_{\hat{A}}=0$, which is equivalent to 
	\begin{equation}
		\begin{split}
			\label{A3}
			\hat{\bm{\beta}}_{\hat{A}}\in\arg\min\frac{1}{2N}||\bm{X}_{\hat{A}}\bm{\beta}_{\hat{A}}-\bm{y}||^2_2.
		\end{split}
	\end{equation}
	Next, we prove $l_\lambda(\hat{\bm{\beta}}+\bm{h})\ge l_\lambda(\hat{\bm{\beta}})$ when $\bm{h}$ is small enough with $||\bm{h}||_\infty\le\sqrt{2\lambda}$. We consider two cases. If $\bm{h}_{\hat{I}}\ne 0$, then
	\begin{equation}
		\begin{split}
			\label{A4}
			l_\lambda(\hat{\bm{\beta}}+\bm{h})-l_\lambda(\hat{\bm{\beta}})&\ge \frac{1}{2N}||\bm{X}\hat{\bm{\beta}}-\bm{y}+\bm{Xh}||^2_2-\frac{1}{2N}||\bm{X}\hat{\bm{\beta}}-\bm{y}||^2_2+\lambda\\
			&\ge\lambda-|\langle\bm{h},\hat{\bm{d}}\rangle|,
		\end{split}
	\end{equation}
	which is established when $\bm{h}$ is sufficiently small. If $\bm{h}_{\hat{I}}=0$, we can deduce $l_\lambda(\hat{\bm{\beta}}+\bm{h})\ge l_\lambda(\hat{\bm{\beta}})$ by minimizing property of $\hat{\bm{\beta}}_{\hat{A}}$ in (\ref{A3}). Then change the loss function to the distributed form (\ref{A2}) and finally completes the proof of Lemma \ref{lemma:1}.
\end{proof}

\noindent\textbf{Lemma A.1}\enspace Denote $\lbrace\tilde{\bm{\beta}}^{(k)},\tilde{\bm{g}}^{(k)},\tilde{\bm{d}}^{(k)},\tilde A^{(k)}\rbrace$ as the solution sequence of the $k$th iteration of the ESDAR algorithm, then $\tilde{A}^{(k)}=\hat{A}^{(k)}$ holds with high probability as $N\to\infty$.
\begin{proof}
	First, we considering $\tilde{\bm{\beta}}^{(0)}=\hat{\bm{\beta}}^{(0)}=\bm{0}$ in Algorithm 1, then we obtain $\tilde{A}^{(0)}=\hat{A}^{(0)}$. Due to $|\tilde{A}^{(k)}|=|\hat{A}^{(k)}|$, we can obtain $\tilde{A}^{(k)}=\hat{A}^{(k)}$ whenever $\tilde{A}^{(k)}\subseteq\hat{A}^{(k)}$ or $\hat{A}^{(k)}\subseteq\tilde{A}^{(k)}$. When $k=1$, we have 
	\begin{equation}
		\begin{split}
			\label{A5}
			\hat{A}^{(1)}=\lbrace i\in S \mid \sqrt{\hat{g}_i^{(1)}}|\hat{\beta}_i^{(1)}+\tau\hat{d}_i^{(1)}|\ge\sqrt{\hat{\bm{g}}^{(1)}}||\hat{\bm{\beta}}^{(1)}+\tau\hat{\bm{d}}^{(1)}||_{(T)}\rbrace,\\
			\tilde{A}^{(1)}=\lbrace i\in S \mid \sqrt{\tilde{g}_i^{(1)}}|\tilde{\beta}_i^{(1)}+\tau\tilde{d}_i^{(1)}|\ge\sqrt{\tilde{\bm{g}}^{(1)}}||\tilde{\bm{\beta}}^{(1)}+\tau\tilde{\bm{d}}^{(1)}||_{(T)}\rbrace.
		\end{split}
	\end{equation}
	Since $\hat{\bm{\beta}}^{(1)}_{A^{(0)}}=\hat{\bm{\beta}}^{(1)}_{\hat{A}^{(0)}}$, we have $||\tilde{\bm{\beta}}^{(1)}_{A^{(0)}}-\hat{\bm{\beta}}^{(1)}_{A^{(0)}}||^2_2=||\tilde{\bm{\beta}}^{(1)}-\hat{\bm{\beta}}^{(1)}||^2_2=O_p(N^{-1/2})$, which implies that $\tilde{\bm{\beta}}^{(1)}$ will converge to $\hat{\bm{\beta}}^{(1)}$ with probability and the rate of convergence is $N^{-1/2}$. Assume $i\in \hat{A}^{(1)}$, we have
	\begin{equation}
		\begin{split}
			\label{A6}
			\sqrt{\hat{g}_i^{(1)}}|\hat{\beta}_i^{(1)}+\tau\hat{d}_i^{(1)}|\ge\sqrt{\hat{\bm{g}}^{(1)}}||\hat{\bm{\beta}}^{(1)}+\tau\hat{\bm{d}}^{(1)}||_{(T)}.
		\end{split}
	\end{equation}
	
	Next we will prove that $i \in A^{(1)}$. By some simple calculation, we have
	\begin{equation}
		\begin{split}
			\label{A7}
			\sqrt{\tilde{\bm{g}}^{(1)}}\cdot(\tilde{\bm{\beta}}^{(1)}+\tau\tilde{\bm{d}}^{(1)})=\sqrt{\hat{\bm{g}}^{(1)}}\cdot(\hat{\bm{\beta}}^{(1)}+\tau\hat{\bm{d}}^{(1)})+
			\begin{pmatrix}
				\sqrt{\tilde{\bm{g}}^{(1)}_{A^{(0)}}}\cdot\tilde{\bm{\beta}}^{(1)}_{A^{(0)}}-\sqrt{\hat{\bm{g}}^{(1)}_{A^{(0)}}}\cdot\hat{\bm{\beta}}^{(1)}_{A^{(0)}}\\
				\tau\sqrt{\tilde{\bm{g}}^{(1)}_{I^{(0)}}}\cdot\tilde{\bm{d}}^{(1)}_{I^{(0)}}-\tau\sqrt{\hat{\bm{g}}^{(1)}_{I^{(0)}}}\cdot\hat{\bm{d}}^{(1)}_{I^{(0)}}
			\end{pmatrix}.
		\end{split}
	\end{equation}
	Now we define $\hat{\bm{\zeta}}^{(1)}=\sqrt{\hat{\bm{g}}^{(1)}}\cdot(\hat{\bm{\beta}}^{(1)}+\tau\hat{\bm{d}}^{(1)})$ and $\hat{\xi}^{(1)}=\min\limits_{i,j\in \mathcal{F}}\lbrace|\hat{\zeta}^{(1)}_i-\hat{\zeta}^{(1)}_j|,i\ne j\rbrace$.
	
	Combined with the Lagrange mean value theorem, we obtain
	\begin{equation}
		\begin{split}
			&||\sqrt{\tilde{\bm{g}}^{(1)}_{A^{(0)}}}\cdot\tilde{\bm{\beta}}^{(1)}_{A^{(0)}}-\sqrt{\hat{\bm{g}}^{(1)}_{A^{(0)}}}\cdot\hat{\bm{\beta}}^{(1)}_{A^{(0)}}||_\infty\\
			\le&||\sqrt{\tilde{\bm{g}}^{(1)}_{A^{(0)}}}\cdot(\tilde{\bm{\beta}}^{(1)}_{A^{(0)}}-\hat{\bm{\beta}}^{(1)}_{A^{(0)}})||_\infty+||(\sqrt{\tilde{\bm{g}}^{(1)}_{A^{(0)}}}-\sqrt{\hat{\bm{g}}^{(1)}_{A^{(0)}}})\cdot\hat{\bm{\beta}}^{(1)}_{A^{(0)}}||_\infty\\
			\le&||\sqrt{\tilde{\bm{g}}^{(1)}_{A^{(0)}}}||_\infty||\tilde{\bm{\beta}}^{(1)}_{A^{(0)}}-\hat{\bm{\beta}}^{(1)}_{A^{(0)}}||_\infty+||\sqrt{\tilde{\bm{g}}^{(1)}_{A^{(0)}}}-\sqrt{\hat{\bm{g}}^{(1)}_{A^{(0)}}}||_\infty||\hat{\bm{\beta}}^{(1)}_{A^{(0)}}||_\infty\\
			\le&||\sqrt{\tilde{\bm{g}}^{(1)}_{A^{(0)}}}||_\infty||\tilde{\bm{\beta}}^{(1)}_{A^{(0)}}-\hat{\bm{\beta}}^{(1)}_{A^{(0)}}||_\infty+\tilde{C}_1||\hat{\bm{\beta}}^{(1)}_{A^{(0)}}||_\infty||\tilde{\bm{\beta}}^{(1)}_{A^{(0)}}-\hat{\bm{\beta}}^{(1)}_{A^{(0)}}||_\infty\\
			\le&\tilde{C}^{(1)}||\tilde{\bm{\beta}}^{(1)}_{A^{(0)}}-\hat{\bm{\beta}}^{(1)}_{A^{(0)}}||_\infty,
		\end{split}
	\end{equation}
	where $\tilde{C}^{(1)}=||\sqrt{\tilde{\bm{g}}^{(1)}_{A^{(0)}}}||_\infty+\tilde{C}_1||\hat{\bm{\beta}}^{(1)}_{A^{(0)}}||_\infty$. Similarly, we obtain
	\begin{equation}
		\begin{split}\label{A9}
			&\tau||\sqrt{\tilde{\bm{g}}^{(1)}_{I^{(0)}}}\cdot\tilde{\bm{d}}^{(1)}_{I^{(0)}}-\sqrt{\hat{\bm{g}}^{(1)}_{I^{(0)}}}\cdot\hat{\bm{d}}^{(1)}_{I^{(0)}}||_\infty\\
			\le&\tau||\sqrt{\tilde{\bm{g}}^{(1)}_{I^{(0)}}}\cdot(\tilde{\bm{d}}^{(1)}_{I^{(0)}}-\hat{\bm{d}}^{(1)}_{I^{(0)}})||_\infty+\tau||(\sqrt{\tilde{\bm{g}}^{(1)}_{I^{(0)}}}-\sqrt{\hat{\bm{g}}^{(1)}_{I^{(0)}}})\cdot\hat{\bm{d}}^{(1)}_{I^{(0)}}||_\infty\\
			\le&\tau\tilde{C}_2||\sqrt{\tilde{\bm{g}}^{(1)}_{I^{(0)}}}||_\infty||\tilde{\bm{\beta}}^{(1)}_{A^{(0)}}-\hat{\bm{\beta}}^{(1)}_{A^{(0)}}||_\infty+\tau\tilde{C}_3||\hat{\bm{d}}^{(1)}_{I^{(0)}}||_\infty||\tilde{\bm{\beta}}^{(1)}_{A^{(0)}}-\hat{\bm{\beta}}^{(1)}_{A^{(0)}}||_\infty\\
			\le&\tilde{C}^{(2)}||\tilde{\bm{\beta}}^{(1)}_{A^{(0)}}-\hat{\bm{\beta}}^{(1)}_{A^{(0)}}||_\infty,
		\end{split}
	\end{equation}
	where $\tilde{C}^{(2)}=\tau(\tilde{C}_2||\sqrt{\tilde{\bm{g}}^{(1)}_{I^{(0)}}}||_\infty+\tilde{C}_3||\hat{\bm{d}}^{(1)}_{I^{(0)}}||_\infty)$. Let $C^{(1)}=\max\lbrace\tilde{C}^{(1)},\tilde{C}^{(2)}\rbrace$, we have
	$$
	C^{(1)}||\tilde{\bm{\beta}}^{(1)}_{A^{(0)}}-\hat{\bm{\beta}}^{(1)}_{A^{(0)}}||_\infty=C^{(1)}||\tilde{\bm{\beta}}^{(1)}-\hat{\bm{\beta}}^{(1)}||_\infty<<\xi^{(1)}
	$$
	holds with high probability when $N$ is large enough. Suppose that $\tilde{A}^{(k)}=\hat{A}^{(k)}$ for $k\ge 1$. We further obtain
	$$
	C^{(k+1)}||\tilde{\bm{\beta}}^{(k+1)}_{A^{(0)}}-\hat{\bm{\beta}}^{(k+1)}_{A^{(0)}}||_\infty=C^{(k+1)}||\tilde{\bm{\beta}}^{(k+1)}-\hat{\bm{\beta}}^{(k+1)}||_\infty<<\xi^{(k+1)}
	$$
	by repeating the above steps, which implies that $\tilde{A}^{(k+1)}=\hat{A}^{(k+1)}$ holds with high probability. Then we assume that $\sqrt{N}(\tilde{\bm{\beta}}^{(k+1)}_{A^{(k)}}-\hat{\bm{\beta}}^{(k+1)}_{A^{(k)}})\to N(0,\Sigma)$, we have 
	$$
	P\lbrace C^{(k)}||\bm{\beta}^{(k+1)}-\hat{\bm{\beta}}^{(k+1)}||_\infty>\xi^{(k)}\rbrace\le T\exp\lbrace-\frac{N(\xi^{(k)}/C^{(k)})^2}{2\sigma^2}\rbrace\to 0.
	$$
	Hence, we finally obtain
	\begin{equation}
		\begin{split}\label{A10}
			&\prod\limits_{k=1}^{K}P\lbrace C^{(k)}||\tilde{\bm{\beta}}^{(k+1)}-\hat{\bm{\beta}}^{(k+1)}||_\infty\le\xi^{(k)}\rbrace\\
			\ge& \exp\lbrace\sum\limits_{k=1}^{K}\log\lbrace1-T\exp\lbrace-\frac{N(\xi^{(k)}/C^{(k)})^2}{2\sigma^2}\rbrace\rbrace\rbrace\to 1.
		\end{split}
	\end{equation}
	The proof is complete.
\end{proof}

\noindent\textbf{Proof of Theorem \ref{th:5}.}
\begin{proof}
	By the triangle inequality, we can obtain that
	\begin{equation}
		\label{A11}
		||\bm{\beta}^{(k+1)}-\bm{\beta}^*||_2\le||\bm{\beta}^{(k+1)}-\tilde{\bm{\beta}}^{(k+1)}||_2+||\tilde{\bm{\beta}}^{(k+1)}-\bm{\beta}^*||_2.
	\end{equation}
	Hence, we can establish our desired outcomes by placing bounds on the two terms on the right-hand side of equations (\ref{A11}), where the first terms measure the error of the gradient descent method and the second terms measure the error of SDAR, which has been extensively examined in \citet{Huang:2018}. Then by Lemma A.1, we can concluded that $A^{(k)}=\hat{A}^{(k)}$ and $\tilde{\bm{\beta}}^{(k+1)}=\hat{\bm{\beta}}^{(k+1)}$, where $\hat{\bm{\beta}}^{(k+1)}$ is the output of the $(k+1)$-th iteration of the CESDAR. Through some straightforward computations, we obtain
	\begin{equation}
		\label{A12}
		\begin{split}
			||\bm{\beta}^{(k+1)}-\bm{\beta}^*||_2&\le||\bm{\beta}^{(k+1)}-\tilde{\bm{\beta}}^{(k+1)}||_2+||\tilde{\bm{\beta}}^{(k+1)}-\bm{\beta}^*||_2\\
			&\le O_p(N^{-\frac{1}{2}})+b_1\gamma^k||\bm{\beta}^*||_2+b_2h_2(T)\\
			&\le b_1\gamma^k||\bm{\beta}^*||_2+b_2h_2(T).
		\end{split}
	\end{equation}
	At this point, if we incorporate the condition under which (A3) holds, we can further obtain
	\begin{equation}
		\label{A13}
		\begin{split}
			||\bm{\beta}^{(k+1)}-\bm{\beta}^*||_2\le b_1\gamma^k||\bm{\beta}^*||_2+b_2\eta_1,
		\end{split}
	\end{equation}
	where $\eta_1$ is defined in (\ref{eq:13}).
	
	After the aforementioned derivation, through a finite number of iterations, we finally obtain
	\begin{equation}
		\label{A14}
		\begin{split}
			||\bm{\beta}^{(k)}-\bm{\beta}^*||_2\le c\eta_1,
		\end{split}
	\end{equation}
	where
	$k\ge\log_{\frac{1}{\gamma}}\frac{\sqrt{s}||\bm{\beta}^*_{A^*}||_\infty}{\eta_1}$.
	
	The proof is complete.
\end{proof}

\noindent\textbf{Proof of Theorem \ref{th:7}.}
\begin{proof}
	By (\ref{eq:10}),
	\begin{equation}
		\label{A15}
		\begin{split}
			||\bm{\beta}^{(k+1)}-\bm{\beta}^*||_2&\le b_1\gamma^k||\bm{\beta}^*||_2+b_2h_2(T)\\
			&\le b_1h_2(T)+b_2h_2(T),\enspace if\; k\ge\log_{\frac{1}{\gamma}}\frac{\sqrt{s}||\bm{\beta}^*_{A^*}||_\infty}{h_2(T)}.
		\end{split}
	\end{equation}
Suppose $\bm{\beta}^*$ is exactly $s$-sparse and $T\ge s$, we have
	\begin{equation}
		\label{A16}
		\begin{split}
			||\bm{\beta}^*|_{A^*\backslash A^{(k+1)}}||_2&\le \gamma||\bm{\beta}^*|_{A^*\backslash A^{(k)}}||_2+\frac{\gamma}{\theta_{T,T}}\sigma\sqrt{s}\sqrt{2\log(p/\alpha)/N}\\
			&=\frac{\gamma}{\theta_{T,T}}\sigma\sqrt{s}\sqrt{2\log(p/\alpha)/N},
		\end{split}
	\end{equation}
	where the first inequality is derived from the proof of Corollary 8 in \citet{Huang:2018}.
	
	Then $A^{(k+1)}\subseteq A^*$ follows from the assumption that
	$$
	||\bm{\beta}^*_{A^*}||_{\min}\ge\frac{\gamma}{(1-\gamma)\theta_{T,T}\zeta}\sigma\sqrt{s}\sqrt{2\log(p/\alpha)/N}>\frac{\gamma}{\theta_{T,T}}\sigma\sqrt{s}\sqrt{2\log(p/\alpha)/N}.
	$$
	This complete the proof of Theorem \ref{th:7}.
\end{proof}

\noindent\textbf{Proof of Theorem \ref{th:9}.}
\begin{proof}
	Similar to the proof of Theorem \ref{th:5}, by the triangle inequality and the Lemma A.1, we can obtain that
	\begin{equation}
		\begin{split}
			\label{A17}
			||\bm{\beta}^{(k+1)}-\bm{\beta}^*||_\infty&\le||\bm{\beta}^{(k+1)}-\tilde{\bm{\beta}}^{(k+1)}||_\infty+||\tilde{\bm{\beta}}^{(k+1)}-\bm{\beta}^*||_\infty\\
			&\le O_p(N^{-\frac{1}{2}})+\frac{4}{3}\gamma^{k}_\mu||\bm{\beta}^*||_\infty+\frac{4}{3}(\frac{4}{1-\gamma_\mu}+1)h_\infty(T)\\
			&\le \frac{4}{3}\gamma^{k}_\mu||\bm{\beta}^*||_\infty+\frac{4}{3}(\frac{4}{1-\gamma_\mu}+1)h_\infty(T).
		\end{split}
	\end{equation}
	In this case, if (A3) holds simultaneously, we have
	\begin{equation}
		\begin{split}
			\label{A18}
			||\bm{\beta}^{(k+1)}-\bm{\beta}^*||_\infty&\le \frac{4}{3}\gamma^{k}_\mu||\bm{\beta}^*||_\infty+\frac{4}{3}(\frac{4}{1-\gamma_\mu}+1)\eta_2,
		\end{split}
	\end{equation}
	where $\eta_2$ is defined in (\ref{eq:18}).
	
	After multiple iterations, we ultimately obtain
	\begin{equation}
		\begin{split}
			\label{A19}
			||\bm{\beta}^{(k)}-\bm{\beta}^*||_\infty&\le c_\mu\eta_2,
		\end{split}
	\end{equation}
	where $k\ge\log_{\frac{1}{\gamma_\mu}}\frac{4||\bm{\beta}^*_{A^*}||_\infty}{\eta_2}$.
	
	The proof is complete.
\end{proof}

\noindent\textbf{Proof of Theorem \ref{th:11}.}
\begin{proof}
	By (\ref{eq:16}),
	\begin{equation}
		\label{A20}
		\begin{split}
			||\bm{\beta}^{(k+1)}-\bm{\beta}^*||_\infty&\le \frac{4}{3}\gamma^{k}_\mu||\bm{\beta}^*||_\infty+\frac{4}{3}(\frac{4}{1-\gamma_\mu}+1)h_\infty(T)\\
			&\le c_{\mu}h_\infty(T)\quad if\; k\ge\log_{\frac{1}{\gamma_\mu}}\frac{4||\bm{\beta}^*_{A^*}||_\infty}{h_\infty(T)}.
		\end{split}
	\end{equation}
Suppose $\bm{\beta}^*$ is exactly $s$-sparse and $T\ge s$, we have
	\begin{equation}
		\label{A16}
		\begin{split}
			||\bm{\beta}^*|_{A^*\backslash A^{(k+1)}}||_\infty&\le \gamma_\mu||\bm{\beta}^*|_{A^*\backslash A^{(k)}}||_\infty+4\sigma\sqrt{2\log(p/\alpha)/N}\\
			&=4\sigma\sqrt{2\log(p/\alpha)/N},
		\end{split}
	\end{equation}
	where the first inequality is derived from the proof of Corollary 14 in \citet{Huang:2018}.
	
	Then $A^{(k+1)}\subseteq A^*$ follows from the assumption that
	$$ ||\bm{\beta}^*_{A^*}||_{\min}\ge\frac{4}{\zeta(1-\gamma_\mu)}\sigma\sqrt{2\log(p/\alpha)/N}>4\sigma\sqrt{2\log(p/\alpha)/N}.
	$$
	This complete the proof of Theorem \ref{th:11}.
\end{proof}

\end{document}